\documentclass[smallextended]{svjour3}

\usepackage{amssymb}
\usepackage{url}
\usepackage{graphicx}
\usepackage[colorlinks, linkcolor=blue]{hyperref}
\usepackage[hyphenbreaks]{breakurl}
\usepackage{amsmath}
\usepackage{underscore}

\begin{document}

\title{Synergetic Learning Systems: Concept, Architecture, and Algorithms}
\titlerunning{SLS: Concept, Architecture, and Algorithms}
%
\author{Ping Guo
\and
Qian Yin
}
\authorrunning{P. Guo \& Q. Yin}
%
\institute{Ping Guo \at Image Processing \& Pattern Recognition Lab., Beijing Normal University,  Beijing 100875, China\\
\email{pguo@bnu.edu.cn}\\
\and
Qian Yin \at Image Processing \& Pattern Recognition Lab.,  Beijing Normal University,  Beijing 100875, China\\
\email{yinqian@bnu.edu.cn}
}

\date{Received: date / Accepted: date}

\maketitle

%
%
\begin{abstract}
Drawing on the idea that brain development is a Darwinian process of ``evolution + selection'' and the idea that the current state is a local equilibrium state of many bodies with self-organization and evolution processes driven by the temperature and gravity in our universe, in this work, we describe an artificial intelligence system called the ``Synergetic Learning Systems''. The system is composed of two or more subsystems (models, agents or virtual bodies), and it is an open complex giant system. Inspired by natural intelligence, the system achieves intelligent information processing and decision-making in a given environment through cooperative/competitive synergetic learning. The intelligence evolved by the natural law of  ``it is not the strongest of the species that survives, but the one most responsive to change,'' while an artificial intelligence system should adopt the law of ``human selection'' in the evolution process. Therefore, we expect that the proposed system architecture can also be adapted in human-machine synergy or multi-agent synergetic systems. It is also expected that under our design criteria, the proposed system will eventually achieve artificial general intelligence  through long term coevolution.

\end{abstract}


\section{Introduction}
\label{sec1}
During the development of artificial intelligence (AI), scholars of different disciplines have refined their understanding of artificial intelligence, put forward different viewpoints, and produced different academic schools of thought. There are three major schools of thought that have great influence on artificial intelligence research: symbolism, connectionism, and actionism.

\begin{itemize}
\item Symbolism, also known as logisticism, psychologism, or computerism, is based on the assumptions of physical symbology (i.e., symbolic operating systems) and the principle of limited rationality.
\item The main principle of connectionism, also known as bionicism or physiologism, is the connection mechanism and learning algorithm between neural networks.
\item Actionism, also known as the theory of evolutionism or cyberneticism, is based on cybernetics and perceptual-action-based control systems.
\end{itemize}

After two troughs, the development of artificial intelligence benefited from the persistence and continuous efforts of Hinton et al. In 2006, the concept and algorithm of deep belief networks were proposed, which rekindled the passion for neural networks in the field of artificial intelligence. In 2012, Hinton and his student Alex Krizhevsky designed AlexNet based on a Convolutional Neural Network (CNN) and took advantage of the powerful parallel computing power of GPUs in an image competition, which is the forefront of computer intelligent image recognition. The test error rate was 15.3\%, which was much lower than the second test error rate of 26.2\%. In 2015, LeCun, Bengio \& Hinton jointly published a deep study review paper \cite{Deeplearning} in the journal Nature, and neural networks experience a strong resurgence under the name \``deep learning\``. Thanks to the explosive growth of interconnected data, the significant increase in computing power and the development and maturity of deep learning algorithms, we have ushered in the third wave of development since the emergence of artificial intelligence.

With the evolution of time and the expansion of research, deep learning has encountered bottlenecks, and the theory of artificial intelligence has stagnated. Gary Marcus, a professor of psychology at New York University, poured cold water on deep learning. He criticized various problems in deep learning, as detailed in the literature \cite{Marcus2018}.

 Academician Xu Zongben,  the professor   at Xi-An Jiao Tong University, stated that \cite{Xu2019three}:

``{\em It is difficult to design topologies, to anticipate effects, and to explain mechanisms with deep learning. There is no solid mathematical theory to support solving these three problems. Solving these problems is the main focus of deep learning for future research.} ''

Recently, M. Mitchell Waldrop published a review article in the Proceedings of the National Academy of Sciences (PNAS) entitled \cite{Waldrop2019News}, ``News Features: What are the Limitations of Deep Learning?''  In this PNAS feature news, Waldrop briefly describes the history of deep learning and believes that all the glorious benefits of computing power have made artificial intelligence flourish today. However, deep learning has many limitations, including vulnerability to counterattack, low learning efficiency, application instability, lack of common sense, and interpretability. From a computability point of view, more and more people in the field of artificial intelligence research believe that to solve the shortcomings of deep learning, some fundamental new concepts and ideas are needed.

Yann LeCun gave the speech entitled  as ``Learning World Models: the Next Step towards AI'' at the opening ceremony of IJCAI-2018 \cite{LeCun2018jicai}.  LeCun said that the future of the artificial intelligence revolution will  be neither supervised learning, nor will it be pure reinforcement learning, but rather a world model with common sense reasoning and predictive ability. Intuitively, the world model contains general background knowledge about how the world works, the ability to predict the consequences of actions, and the ability to have long-term planning and reasoning. Yann LeCun summarized three learning paradigms, namely, reinforcement learning, supervised learning and self-supervised learning, and believes that self-supervised learning (formerly known as predictive learning) is a potential research direction to realize the world model. At the end of the lecture, Yann LeCun summarized the mutual drive and promotion between technology and science, such as telescopes and optics, steam engines and thermodynamics, or computers and computer science. He also raised several questions:
{\em
\begin{enumerate}
\item What is the equivalent of ``thermodynamics'' in intelligence?
\item Are there underlying principles behind artificial intelligence and natural intelligence?
\item  Are there simple principles behind learning?
\item Is the brain a collection of a large number of ``hacks'' that evolved?
\end{enumerate}
}

As there are many schools of thought regarding basic research on artificial intelligence, it is difficult to construct a unified theory to solve these questions. However, we believe that computational intelligence (CI) is a new stage in artificial intelligence  development of artificial intelligence. CI, a nature-inspired intelligence, and a paradigm that has the potential to solve most problems. CI is initialized  from the phenomena and laws of physics, chemistry, mathematics, biology, psychology, physiology, neuroscience and computer science, and integrates the three artificial intelligence schools of thought to form an organic whole. The system formed by the integration of multiple disciplines and technologies can realize complementary advantages, which will be more effective than a single discipline or technology and can achieve greater results. Therefore, in order to solve the shortcomings of deep learning, we propose the use of the cognitive neuroscience mechanism and mathematical tools in machine learning to develop a new generation of artificial intelligence. Based on computational intelligence, we should develop a Synergetic Learning Systems (SLS) \cite{Synergetic2019} to establish a theoretical foundation for intelligent ``thermodynamics.''

For a more detailed analysis of the current status of artificial intelligence basic research and development trends, please refer to the literature \cite{ChinaXiv2019}.

The structure of Part I (this paper) is as follows: the methodology for developing a Synergetic Learning Systems is given in section 2, section 3 introduces the basic concept of the Synergetic Learning Systems, section 4 describes the architecture of the Synergetic Learning Systems, section 5 lists the relevant optimization algorithms, and the last section summarizes and analyzes the future direction of the Synergetic Learning Systems.

\section{Methodologies}
\label{sec2}
We know that a methodology is a theoretical system aimed at solving problems, usually involving the elaboration of problem stages, tasks, tools, and methodological techniques.

We believe that the artificial intelligence system needs to be analyzed and studied systematically at multiple scales, levels, and perspectives. The so-called multi-scale refers to the study of artificial intelligence systems from the macro scale, micro-scale, and mesoscopic scale. Macroscopic, microscopic and mesoscopic are are our ``Three perspectives'' \cite{method-SLS}.

In the theme report for the Second China System Science Conference, ``How does the brain work in the whole?'' Academician Guo Aike,  a neuroscience and biophysicists in China,  mentioned: ``The brain function linkage map should be drawn from the macro-brain scale, the mesoscopic neural network scale, and the micro synapse scale, and more scales can be considered.'' \cite{CSSC08}

We can also draw on the research results of basic disciplines, such as the evolution of the universe on a macroscopic scale. The main physical parameters for the evolution of the universe are temperature and gravity. According to the Big Bang Theory, the initial temperature was very high, and the current galaxy structure evolved due to cooling and gravity. Therefore, the temperature of a system is a very important basic quantity. In the evolution of the cooperative learning system, the simulated annealing algorithm can be considered in order to solve the combinatorial optimization problem.

To analyze and study the artificial intelligence system systematically is to adopt the methodology of system science and use the viewpoint of the system to understand and grasp the essence and movement development of artificial intelligence. Our proposed Synergetic Learning Systems is inspired by natural intelligence and is based on the cognitive neuroscience mechanism and computational intelligence, integrating multidisciplinary knowledge and adopting the mode of complex system thinking. The Synergetic Learning Systems is based on the concept of systems science. The use of the term ``collaboration'' is influenced by the idea of ``synergetic learning'' \cite{Hermann10}\cite{Hermann11}. However, SLS includes cooperative synergetic learning and competitive synergetic learning among various elements, as well as concepts and methods, such as system evolution and evolutionary game.

In the field of neural network research, Hinton's Boltzmann machine, the Helmholtz machine, and the restricted Boltzmann machine draw on the concepts and research methods of statistical physics. Therefore, we also need to look at the problem by using physical thinking during the research process. Statistical mechanics looks at the essence through phenomena, in which the phenomena is the observed data and essence is the law. In statistical mechanics, probability distribution, mathematical models and other tools are used to systematically quantify and analyze the general laws and randomness behind observed data.

For a more detailed discussion, please refer to \cite{method-SLS}.

\section{Basic Concept}

The Synergetic Learning Systems we proposed is an information processing system, which is equivalent to an intelligent ``thermodynamic system.'' As we know, neural networks process information to achieve intelligent information processing and decision making in a given environment. The rule of ``natural selection, survival of the fittest'' followed by nature should adopt the ``human selection'' rule when the artificial intelligence system evolves. We believe that this law is the principle of free energy. Nature likes to find a physical system with minimum  free energy, so free energy can also be utilized as an objective function of system evolution.

The concept of free energy comes from statistical physics. It refers to the part of the system that can be converted into external work in a certain thermodynamic process. In a particular thermodynamic process, the ``useful energy'' of the system's external output can be divided into Helmholtz free energy and Gibbs free energy.

In  NIPS'93 , Hinton et al. \cite{Hinton1994Autoencoders12} established relationship among the auto-encoder and minimum description length (MDL) principle with the Helmholtz free energy, and also transformed the auto-encoder into a restricted Boltzmann machine. Hinton borrowed  concepts from statistical physics and explained it as the deep belief network. Based on the viewpoint of statistical physics and the relationship among  free energy, internal energy and entropy in the canonical ensemble, the interpretability problem of the model can be solved. In statistical physics, ensembles represent a collection of a large number of possible states of a system under certain conditions. In the canonical  ensemble, the relationship among free energy $F$, internal energy $E$, entropy $S$, and temperature $T$ of the state function is $F = E - TS$, and the relationship between the free energy and the partition function $Z$ is:

\begin{eqnarray*}
F =k_{B}T\ln Z
\end{eqnarray*}

where $k_{B}$ is the Boltzmann constant. Entropy is a linear combination of free energy $F$, temperature $T$ and average internal energy $\langle E \rangle$, $S=(\langle E \rangle - F) / T$. The partition function is 
$ Z=\sum_{i}{ \Omega }(E_{i})e^{-\beta E_{i}} $ , 
where ${ \Omega} (E)$ is the level of degeneracy and $\beta = 1/(k_{B}T)$. In a canonical  ensemble, the probability distribution function is the Boltzmann distribution, $p_{i}=\frac{1}{Z}\exp ^{-\beta E_{i} }$. Therefore, as long as the free energy of the ensembles is defined and the combined network model is optimized by the principle of least action, the desired neural network systems can be obtained.

The energy-based learning algorithm \cite{Lecun2006A13} should also be derived from statistical physics. Hinton also made the neuroscientist Karl Friston accept the idea that the best way to explore the brain is to think of it as a Bayesian probability machine. In 2010, Friston published a paper titled, ``The free-energy principle: a unified brain theory?'' in Nature Reviews Neuroscience, explaining the brain's operating mechanism using the principle of free energy \cite{Friston2010}. From the work of Hinton et al. and Friston, we are convinced that the study of the ``intelligent  thermodynamics'' system based on the principle of free energy undoubtedly contributed to the inspiration and success of the Synergetic Learning Systems theory.

To break through those dilemmas of topology design of poor deep learning network structure,  difficulty of predicting the effects and explainability of  the network mechanism, we propose to build a Synergetic Learning Systems to establish a ``grand unified theory'' of intelligence.

Drawing on system theory to study SLS, the most fundamental Synergetic Learning Systems we designed has two subsystems (or models): the system reduction model (discriminative model) and the system evolution model (generative model). The evolution of the system is described by a differential dynamic system.

At the IJCAI 2018, one of Yann LeCun's questions was, ``is there a simple rule behind learning?''  We think there should be a simple rule. It is well known that there are simple and elegant principle in physics, which is the {\bf principle of least action}.

A mechanical system, using the result of variation of the Hamiltonian by the principle of least action, can derive the Lagrangian equation describing the mechanical system. If we define the total effect of the system as equal to the sum of the actions of the gravitational field and the material field, Einstein's  general relativity equation can be derived also according to the principle of least action.

Our proposed Synergetic Learning Systems is an information processing system based on the principle of free energy. Therefore, we assume  that in an SLS, the amount of action in the system is equal to the free energy.  The principle of free energy in the SLS is, in particular, equivalent to the principle of least action, and free energy is equivalent to the Hamiltonian of the mechanical system. Therefore, our proposition is:

Free energy == Hamiltonian;

Principle of free energy == principle of least action.

Therefore, for a given environment (data), as long as the ``Hamiltonian'' of the neural network system is defined, the self-organization and evolution of the neural network structure can be systematically studied through multi-view and multi-scale dynamics equations.
This gives us an important concept also, that is,  {\bf principle of least action is the first principle for artificial intelligence.}

The dynamics of the SLS should be described by differential dynamic equations. What kind of equation is this differential dynamic equation?
We know that in the field of chemical research, the ``free energy'' of a one-component system can be represented by a variational form as a restricted evolution equation:\cite{wikiRDE2016}

\begin{eqnarray}
\frac{\partial \phi }{\partial t}=-\frac{\delta \mathcal{F}}{\delta \phi}
\end{eqnarray}

``Free energy''  $\mathcal{F}$ is given by:

\begin{eqnarray}
\mathcal{F}=\int_{-\infty }^{\infty }\left [ \frac{D}{2} \left ( \frac{\partial \phi}{\partial x} \right )^{2}-V\left ( \phi \right )\right ]\mathrm{d} x
\end{eqnarray}

In the above formula, $V(\phi)$ is the chemical potential and $R(\phi)={\mathrm{d} V(\phi))}/{\mathrm{d} \phi}$. Therefore, we have
\begin{equation}
 \frac{\partial \phi}{\partial t}=D\frac{\partial^2 \phi}{\partial^2 x}+R(\phi).
\end{equation}

If generalizing the equation to a multi-component systems, we can obtain a variation of the ``free energy'' $\mathcal{F}$:

\begin{eqnarray}
\frac{\partial \psi }{\partial t}=\nabla \cdot (D(\Psi )\nabla\Psi )+\nabla(f(\Psi))+g(\Psi)
\end{eqnarray}

in which $\psi (x,\theta ,t)=\left \{ \phi _{1},\phi _{2},\cdots,\phi _{n}  \right \}$, $D(\psi)$ is the diffusion matrix, and $\nabla \psi=\left ( \frac{\partial \psi}{\partial x_{1}}\vec{e_{1}}+\frac{\partial \psi}{\partial x_{2}}\vec{e_{2}}+\cdots \cdots+ \frac{\partial \psi}{\partial x_{n}}\vec{e_{n}}\right )$.  $\nabla f(\psi)$ is called convection vector and $g(\psi)$ is  reaction vector\cite{Qixiao16}.

Therefore, the free energy of the systems is equal to the amount of action, and the variation algorithm is applied to action according to the principle of least action (the principle of free energy), and the reaction-diffusion equation is derived! Therefore, the dynamics of the SLS is described by the reaction-diffusion equation. In statistical physics, the dissipative systems theory is a theoretical description of the self-organization of non-equilibrium systems, and  the reaction-diffusion equation is utilized to model this systems also \cite{Willems1972}.

Below we describe the architecture of the Synergetic Learning Systems.

\section{System Architecture}
The unity of  structure and function is one of the basic concepts in biology. The brain has a complex neural network structure. Therefore, a unified architecture is very important, and it is unified with the function of the system.

Aristotle, a famous ancient Greek philosopher, proposed that ``the whole is greater than the sum of the parts,'' which is an ancient, simple overall view and a basic principle of modern systems theory. In accordance with this basic principle, we have designed a SLS that consists of two or more subsystems. The SLS should not  be a simple isolated system. It contains at least two subsystems, in this way  the performance of the systems may demonstrate the  the whole is greater than the sum the parts, otherwise, only one part  summed still is one part. It is well known that the brain is an open complex giant system,  if we intend to simulate the brain structures  with neural networks, the SLS should also be an open complex giant system.

According to Dr. Hsue-Shen Tsien's category method \cite{XueSen1990}, if there are many kinds of subsystems and a hierarchical structure in a systems, the relationship between them is very complicated, thus giving rise to a complex giant system. If this system is also open, it is called an open complex giant system. Openness at here refers to the exchange of energy, information or matter between the systems and the outside world. To be more precise:  1. the systems and its subsystems have various exchanges of information with the outside world;  2. each subsystems in the systems acquires knowledge through learning.

Academician Guo Aike elaborated on the working principle of the brain and the roots of his intelligence: ``How does the human brain work as a whole?   `The Tao produced One, One produced Two, Two produced Three, Three is All things.'  My initial understanding is the result of a multi-module synergetic operation; I believe that the function of the brain is the result of the collaboration of thousands of subsystems with different specialized skills, which is the result of the entanglement combination of millions of years of evolution'' \cite{CSSC08}. Therefore, referring to the neurocognitive mechanism, the overall working state of the SLS is also the result of the coordinated operation of multiple subsystems (multi-modules).

The visual organ of \textit{Drosophila} consists of more than 750 monoculars. William Bialek, a theoretical physicist at Princeton University, has shown that these eyes work together to create a visual system that enables highly accurate calculations \cite{Haykin2011nnnlm}. This systems illustrates the synergy of individuals in a complex systems, and it is one of the biological basis for our proposed SLS.

Figure 1 is a schematic diagram of the architecture of our Synergetic Learning Systems. In this SLS, subsystems can have multiple structures and layers. The system is flexible and scalable and has complex interrelationships between subsystems.

\begin{figure}
\centering
\includegraphics[width=.8\linewidth]{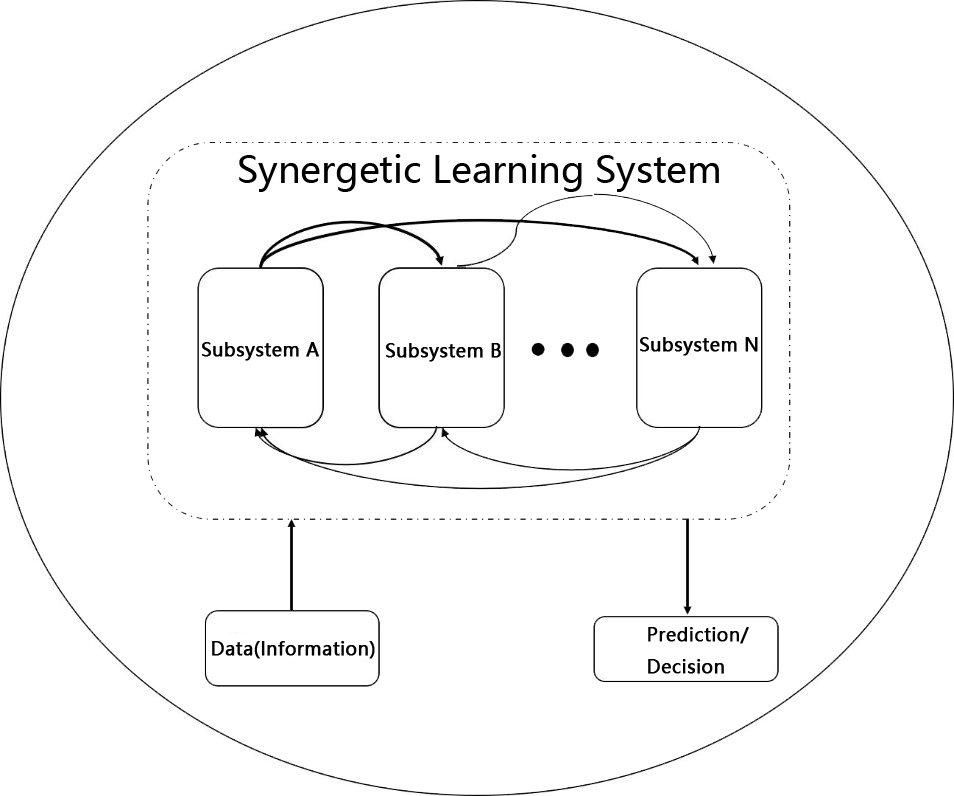}
\caption{Schematic diagram of the architecture of our Synergetic Learning Systems.}
\label{fig:frog}
\end{figure}

\subsection{Multi-Agent Systems}

If each subsystem is a peer-to-peer model, and each subsystem is an agent, the Synergetic Learning Systems will be a Multi-Agent Systems (MAS).

A MAS is a collection of multiple agents. Its goal is to transform large, complex systems into small, coordinated, and manageable systems that can easily communicate with each other. Therefore, we can understand specific examples of the application of the divide-and-conquer strategy.

The MAS is a coordination system, and each agent solves large-scale complex problems by coordinating with each other. The MAS is also an integrated systems, which uses information integration technology to integrate the information from each subsystem to complete the integration of complex systems.

In a MAS, each agent communicates and coordinates with each other and solves problems in a parallel way, thereby this will effectively improve its ability to solve complex problems. Multi-Agent Systems are suitable for complex, open distributed systems. They deal with the task through the cooperation of the agent. The key to realizing the MAS is the communication and coordination between these agents, that is, the synergetics. After the data are given, the processing of building the MAS is the processing of synergetic learning.

We know that data is the manifestation and information carrier. Therefore, a specific SLS relies on a given data set, that is, it is a data-driven modeling process. As the systems evolves, data-driven single-wheel drive will evolve into the two-wheel drive comprised of data and models.

We believe that the MAS and swarm intelligence are highly similar, but the scope of swarm intelligence  may be wider than that of the MAS. The SLS focuses on how each agent works with others. Each subsystem can be a complex neural network system, and much more attention to systems's  integrity is paid  compared with  that of the MAS or swarm intelligence systems.

One of special examples is that if a hybrid expert systems architecture is used, the gate network can be used to collaborate the opinions of various expert networks. The gate network can be designed to utilize simple voting or weighted voting mechanics. In addition, the gate network can be connected to various expert networks to coordinate the input and output of the expert networks according to given task, which is the servo mechanism in the synergetic learning. The gate network controls the various expert networks, when from viewpoint of the stacked generalization,  we can regard the gate network module as a meta learner.

\subsection{Two-body system}

Based on reductionism, a complex systems  consists of multiple simple systems. We should go from simple to complex when tackling problems. The nonliving system usually obeys the second law of thermodynamics, as a system always spontaneously tends toward equilibrium and disorder, and the entropy of the system achieves a maximum  value eventually. The system spontaneously changes from order to disorder, while disorder does not spontaneously change to order, which is due to the irreversibility of the system and the stability of the equilibrium state. However, the life system is the opposite. Biological evolution and social development are always more orderly:  from simple to complex and from low to high order. Such systems are capable of spontaneously forming an orderly, stable structure. Life evolves from a simple structure to a complex form also,  therefore, we can start to construct a system from a system with simple structures and gradually evolve into complex systems. The two-body system is the most fundamental  SLS. In the AI field, there are a lot of examples about two subsystems. On the other hand, when  dealing with many body problems in statistical physics, we may take  mean field approximation. With this approximation,  the many body problem can be simplified into the two-body problem.  Although the degree of approximation is related to a specific problem, it is a proven method of solving complicated problems effectively.

\subsubsection{Universal two-body systems concept}
Any one systems can be divided into two subsystems. In the AI field there are many dual systems, for example, computer graphics, in which the GPUs are widely used, are dual counterpart of  images. A Chinese to English translation system has a dual translation system for English to Chinese. These two systems can be considered as subsystems to form a large systems, which can be viewed as dual learning / Synergetic Learning Systems. In the probabilistic statistical model, the Generative Adversarial Network (GAN) and the Variational Auto-Encoder (VAE) can be considered as Synergetic Learning Systems with two subsystems also.

\subsubsection{Simple two-body systems}

The simplest two-models system is the Autoencoder. Here we regard the encoder is the system A,  the decoder is the system B, and the two parts are combined into a simple SLS.

In this simple SLS, the cooperative manner is sequential, and the input of the decoder depends on the output of the encoder. We can interpret  that these two subsystems are enslaved by the loss function.  Minimizing the loss function (reconstruction error) is one way to achieve synergetic learning.

The restricted Boltzmann machine can also be considered a simple SLS.
A detailed discussion will take place in another  work \cite{Synergetic2019b}, the part II of the SLS  is primarily concerned with the interpretability of neural network systems based on statistical physics. For the marginal distribution of the joint distribution $p(v, h)$, it is understood that $p(v)$ is a subsystem and $p(h)$ is another subsystem. In fact, many energy-based models are called Boltzmann machines in the past. The original Boltzmann machine was composed of two types of models: models with and without latent variables. What is now called the Boltzmann machine is a model with latent variables \cite{Ian18}.

\section{Algorithms}

Under the framework of the free-energy principle, the SLS can be trained  with  energy-based learning algorithms. The state of a physical system can be studied when  ``action'' is defined, and the maximum / minimum value is obtained by the variational method based on the principle of least action \cite{gray2004progress}. In the SLS, we need to define the free energy of the systems. After adopting statistical physics thinking,  the mathematical tool to describe the model is probability and statistics. Hopfield neural networks were early energy-based models \cite{Hopfield40} \cite{Hopfield41}. In an energy-based learning model, negative variational free energy is also known as the Evidence Low BOund (ELBO)\cite{Springer43}\cite{Fox2012}\cite{Beal2003}. To the estimate the probability density function, we can adopt the Expectation-Maximization algorithm \cite{Ian18}, the Markov Chain Monte Carlo (MCMC) algorithm \cite{Bishop2006}and the variational inference algorithm \cite{Jordan1998}\cite{DavidVIreview2017}.

Here we give some examples in the equilibrium state (thermal equilibrium):

\subsection{Variational inference algorithm}

In the paper of Yann LeCun et. al. titled ``A Tutorial on Energy-Based Learning'', the variational free energy is defined as follows\cite{Lecun2006A13}:

\begin{eqnarray*}
L_{nll}(W,Y^{i},X^{i})=E(W,Y^{i},X^{i})+\mathcal{F}_{\beta }(W,\mathcal{Y},X^{i})
\end{eqnarray*}
Where $\mathcal{F}$ is the  \textit{free erengy} of the ensemble $\{E(W,y,X^{i}),y \in \mathcal{Y}\}$:

\begin{eqnarray*}
\mathcal{F}_{\beta }(W,\mathcal{Y},X^{i})=\frac{1}{\gamma }\log\left ( \int_{y\in \mathcal{Y}} \exp \left ( -\beta E(W,y,X^{i}) \right )\right )
\end{eqnarray*}

In the paper titled ``Energy-based Generative Adversarial Network'' \cite{DBLP:46},  discriminator is described as an energy function (negative evaluation function).That is, the smaller the function is, the truer the data. The auto-encoder AE is used as a discriminator (energy function). The energy function is defined as the error function of the discriminator:

\begin{eqnarray*}
L_{D}(x,z)=D(x)+max\left ( 0,m-D(G(z)) \right )
\end{eqnarray*}

According to the Boltzmann distribution, $p_{i}(\Theta )=1/Ze^{-\beta E_{i}}=1/Ze^{-\beta L_{D}}$ and the distribution function $Z(\Theta )=\sum_{i}L_{D}(x_{i},z_{i},\Theta )$. The free energy can be expressed as follows:

\begin{eqnarray*}
\mathcal{F}=k_{B}T\ln Z(\Theta )
\end{eqnarray*}

From these equations, we can see that you can calculate the free energy just by finding the partition function. Bengio et al. turned the problem into an estimating  a probability distribution problem\cite{Kim2016}\cite{Kumar2019}. Therefore, the key issue of the problem is to estimate the probability density distribution function, and one of the algorithms is variational inference algorithm. Some introduction about the variational inference algorithm can be found at literatures \cite{Springer43} \cite{Beal2003}.

In order to utilize  variational inference (variational Bayes) algorithm to solve the SLS learning problem, we need to define a certain environment, in other words, to  assume some conditions. Suppose we design a form of a complete Bayesian model,  in which all parameters are given in the prior distribution. The model has both parameters and potential variables.  when $Z$ is used to represent the set of all potential variables and parameters, we use $X$ to represent the set of all the observed variables. For example, we might have a set of $N$ independent, identically distributed (i.i.d) data, where $X = \{x_1, \dots, x_N\}$ and $Z = \{z_1, \dots, z_N\}$.

One of our models represents the joint distribution $p(X, Z)$, and our goal is to find an approximation of the posterior distribution $p(Z\vert X)$ and model evidence $p(X)$.

In general, the form of posterior probability is very complicated and difficult to be obtained, so we hope to approximate $p(Z|X)$ with a relatively simple and easily understood model $q(Z|X)$, namely, $p(Z|X) \thickapprox q(Z|X)$. Another model is described by $q(X,Z)$, and   $q(Z)=\int q(X,Z){\mathrm{d} x}$.

Mean filed approximation (Factorized distributions): The local interaction between individuals in the system can produce a relatively stable behavior at the macro level, therefore, we can make the  posterior is independence hypotheses. That is,

\begin{eqnarray*}
\forall i,  p(Z|X) &=& p(Z_{i}|X)p(Z_{-i}|X),\\
q(Z) &=& \prod_{i=1}^{M}q_{i}(Z_{i})
\end{eqnarray*}

Since logarithmic evidence, $\log p(X)$, is fixed by the corresponding $q$, in order to minimize the Kullback-Leible (KL) divergence, only $L(q)$ should be maximized. By selecting an appropriate $q$, $L(q)$ is easily calculated and evaluated. In this way, the approximate analytical expression of the posterior $p(Z|X)$ and the lower bound of the log evidence can be obtained, which is also called the variational free energy:

\begin{eqnarray}
L(Q) &=&\sum_{z}q(Z|X)\log p(Z,X) - \sum_{z}q(Z|X)\log q(Z|X)\nonumber \\
&=& \mathbb{E}_q [\log p(Z,X)]+H(q).
\end{eqnarray}

For this equation, the first term on the right-hand side is defined as the energy, and the second term is the Shannon entropy. A more detailed discussion of problem solving can be found in chapter 10 of Bishop's book \cite{Bishop2006}.

\subsection{Approximate  Synergetic Learning Algorithms}
\label{sec-ASLA}
 
 As discussed above,  we can  consider minimizing  a Helmholtz  Free Energy,   this is equivalent to minimizing  the expected log likelihood, under the model
$$
\min\mathbb{E}_{q(z)}[\log p(\mathbf{x})]\; .
 $$

Because the ELBO is equivalent to negative variational free
energy, we maximize the ELBO is the same with minimizing
variational free energy.  
Of cause,  we can also start by just minimizing the KL divergence between the models.  It is notice that the SLS we discussed above
  is consist of two sub-systems (models, or  agents),  one model is expressed with $p({\bf x}, {\bf z} ) $,  and the other is described with $q({\bf x}, {\bf z} )$.  
Therefore,  to implement synergetic  learning for this two-models SLS,   it results to following optimization problem.

\begin{equation}
\label{eqKL}
\min\limits_{\Theta} KL [ q({\bf x}, {\bf z} ) \|p({\bf x}, {\bf z} ) ] .
\end{equation}

Where $\Theta$ is a parameter group we try to learn.  In most cases this is intractable, but if we sophisticated  design the proper models, this will  be tractable. The key issue is that we can use gradient descent optimization or pseudo-inverse learning (PIL) algorithm to train the neural network model.  Consequently,  we developed the 
 approximate  synergetic learning (ASL) algorithm, which is based on our previous work \cite{Guo2002} \cite{Guo2003reg} \cite{Guo2008}, to tackle the complicated variational inference computation problems.

In our ASL algorithm,  the models are designed as follows:

$p(\mathbf{x})$ is a Gaussian mixture distribution, and  $q({\bf x} ) $ is a nonparametric kernel estimation.

\begin{equation}
p(\mathbf{x}) = \sum _{k=1}^{K} \alpha _k \mathcal{N} (\mathbf{x}| \mathbf{\mu} _k , \mathbf{\Sigma }_k). 
\end{equation}
 
\begin{equation}
q(\mathbf{x}) = K (\mathbf{x}, h), 
\end{equation}
where $h$ is a kernel parameter.

\begin{eqnarray}
KL [ q({\bf x}, {\bf z} ) \|p({\bf x}, {\bf z} ) ] & = &\int q({\bf x}, {\bf z} ) \ln \frac{q({\bf x}, {\bf z} )}{p({\bf x}, {\bf z} )} dx dz\\
& = & \int  q( {\bf z} |{\bf x}) q({\bf x}) \ln \frac{q( {\bf z} |{\bf x}) q({\bf x})}{p({\bf x} |{\bf z} ) p({\bf z} ) } dx dz\nonumber 
\label{eqKL1}
\end{eqnarray}

Let 

\begin{equation}
q( {\bf z} |{\bf x})  =  \frac{\alpha _k \mathcal{N} (\mathbf{x}| \mathbf{\mu} _k , \mathbf{\Sigma }_k)}{p({\bf x}, {\bf \Theta} )} \nonumber
\label{eqKL2}
\end{equation}

\begin{equation}
p({\bf x},  \mathbf{z})  =  p({\bf x}, {\bf \Theta} )  =\sum _{k=1}^{K} \alpha _k \mathcal{N} (\mathbf{x}| \mathbf{\mu} _k , \mathbf{\Sigma }_k),
\label{eqKL3}
\end{equation}

With this definition,  $ q( {\bf z} |{\bf x}) $ is  posterior probability estimation in $E-step$ of EM algorithm. This means

 \begin{eqnarray}
\int q( {\bf z} |{\bf x}) q({\bf x} ) \ln  p({\bf x}, {\bf \Theta} ) d \mathbf{x}  d\mathbf{z} &= & \int q({\bf x} ) \ln  p({\bf x}, {\bf \Theta} ) d \mathbf{x}  \nonumber  \\
  \int  q( {\bf z} |{\bf x}) q( {\bf x})  \ln q( {\bf x}) d \mathbf{x} d\mathbf{z} &=& \int  q( {\bf x})  \ln q( {\bf x}) d \mathbf{x} \nonumber 
\label{eqKL4}
\end{eqnarray}
(Normalized in hidden space, $\int q( {\bf z} |{\bf x})  d\mathbf{z}=1$.)

Then 
\begin{eqnarray}
KL [ q|p ]  = & - & \int q({\bf x} ) \ln  p({\bf x}, {\bf \Theta} ) d \mathbf{x}  \\
& + & \int  q( {\bf x})  \ln q( {\bf x}) d \mathbf{x} \nonumber 
\label{eqKL5}
\end{eqnarray}

\subsubsection{Cluster Number Selection}
The details of this work are discussed  in Ref. \cite{Guo2002}.

With data set $D=\{\mathbf{x}_i \}_{i=1}^N$, we intend to cluster the data into several clusters. 

Now we use Gaussian kernel density for $q({\bf x} ) $ ,

\begin{equation}
q(\mathbf{x})  = \frac{1} {N}\sum _{i=1}^{N}  \mathcal{N} (\mathbf{x}| \mathbf{x} _i ,  h^2 \mathbf{I}_d).
\end{equation}
The hyper-parameter $h$ play the key rule in cluster number selection problem, if it estimated with gradient descent approach, it will approach
to zero eventually.  With minimizing KL divergence (Free energy ), we obtain
a new equation for estimating smoothing parameter $h$.

Let 

\begin{equation}
g(\mathbf{x}, \mathbf{\Theta })  = - \ln p({\bf x}, {\bf \Theta} ),
\end{equation}
and use Taylor expansion for $g(\mathbf{x}, \mathbf{\Theta })$ at $\mathbf{x}= \mathbf{x} _i$.
When $h$ is small, we can omit the higher order terms and only keep the first-order term.

\begin{equation}
h^2   =  \frac{\frac{1}{2N} \sum _{j=1}^{N} [ q(\mathbf{x}_j) -1 ] \ln q(\mathbf{x}_j) }{\mathbf {J}_r(\mathbf{x} _i , \mathbf{\Theta}) },
\end{equation}
where
\begin{equation}
\mathbf{J}_r(\mathbf{x} _i , \mathbf{\Theta}) =  \frac{1}{2N} \sum _{i=1}^{N} \left \| \sum _{k=1}^{K}q (\mathbf{z}_k|\mathbf{x}_i )(\mathbf{x}_i - \mathbf{m}_k ) ^T  \mathbf{\Sigma }_k^{-1}\right \|^2.
\end{equation}

Now $\mathbf{z}_k =k$,  stands for Gaussian component label.

\subsubsection {Regularization Parameter Estimation for FNN}

The details of this work are discussed  in Ref. \cite{Guo2003reg}.

Given data set $D=\{\mathbf{x} _i , \mathbf{z} _i  \}_{i=1}^N$,  for supervised learning, $\mathbf{z} _i  $ can be output label, or samples drawn from regressed function.

Now the joint distribution $ q(\mathbf{x}, \mathbf{z}) $   in this work is designed as

\begin{equation}
\label{q-dis-xy-ind}
q(\mathbf{x}, \mathbf{z})  =  \frac{1}{2N} \sum _{\mathbf{x}_i, \mathbf{z}_i \in D} K_{h_x}(\mathbf{x}-\mathbf{x}_i) K_{h_z}(\mathbf{z}-\mathbf{z}_i) .
\end{equation}
where  the kernel density function used the most is Gaussian kernel,

\begin{equation}
K_{h}(\mathbf{r})  =  \mathcal{N} (\mathbf{r}| 0, h\mathbf{I}_d) =\frac{1}{2\pi h^{d/2}} \exp \left\{ - \frac {\|  \mathbf{r}\| ^2}{2h}   \right \}.
\end{equation}

This model is under a very strong assumption ($\mathbf{x}, \mathbf{z}$ are statistically independent ).

When error  is Gaussian \cite{Bishop1996},

\begin{equation}
t_k = h_k(x) +\epsilon _k
\end{equation}

\begin{equation}
p(\epsilon _k)= \frac{1}{2\pi \sigma ^{1/2}} \exp \left ( - \frac {\epsilon _k ^2}{2\sigma ^2}  \right ).
\end{equation}

In our work \cite{Guo2003reg}, the following designed model is considered:

\begin{equation}
p(\mathbf{z}|\mathbf{x}, \mathbf{\Theta}) = p(\mathbf{z}|f(\mathbf{x}, \mathbf{\Theta}))
\label{equationmapping}
\end{equation}
where $f(\mathbf{x}, \mathbf{\Theta})$  is a function of  input variable $\mathbf{x}$ and parameter $\mathbf{\Theta}$.

Also, we design 

 $p(\mathbf{z}|f(\mathbf{x}, \mathbf{\Theta})) = \mathcal{N} (\mathbf{z}| g(\mathbf{x}, \mathbf{\Theta}), \sigma ^2 \mathbf{I}_d)  $,
 {and $ g(\mathbf{x}, \mathbf{\Theta})$ is a neural network mapping function},  for example, single hidden layer feedforward neural networks, or deep neural networks, or any other network architecture. 

\begin{eqnarray}
\mathbf{z} &=& g(\mathbf{x}, \mathbf{\Theta}) +\mathbf{\epsilon} \nonumber \\
\| \mathbf{\epsilon}\|^2 &=&  \| \mathbf{z}- g(\mathbf{x}, \mathbf{\Theta}) \|^2. 
\label{nnmapping}
\end{eqnarray}

In our method, the Taylor expansion approximation is used, this method can be considered as  ``Local quadratic approximation.''
 {(Eq. (17) in Ref. \cite{Guo2003reg}).}

We derived the loss function as follows:

\begin{eqnarray}
\label{reg-eqn}
 loss (\Theta, h) &=& \frac{1}{2N\sigma ^2} \sum _{i=1}^ N \Bigg\{ \| \mathbf{z}_i - g (\mathbf{x}_i, \mathbf{\Theta}) \|^2   \nonumber \\ 
 & + & h \left[   \| g^{\prime} (\mathbf{x}_i, \mathbf{\Theta}) \| ^2  \right ] \\
 & - &  h   \| [ \mathbf{z}_i - g(\mathbf{x}_i, \mathbf{\Theta}) ] g^{\prime \prime} (\mathbf{x}_i, \mathbf{\Theta})   \| \Bigg \} \nonumber .
\end{eqnarray}

The first term is the traditional sum-square-error function, the second term is Jacobin regularization term, and the third term is Hessian regularization term. $h$ is the regularization parameter, which can be estimated with following formula (Eq.(50) in \cite{Guo2003reg}, without considering Hessian term.)
Also assume that prior $p(x)$ is a uniformly distributed function and regard it as $h$ independent,

\begin{equation}
\label{reg-para-est}
h \approx d^2 [1+(d-1)^2] \frac{\sum_{i=1}^{N} \| {\bf z}_i -{\bf g}({\bf x} _i, \Theta)  \|^2 }
{\sum_{i=1}^{N} \|{\bf g}^{\prime} ({\bf x} _i, \Theta) \|^2 }.
\end{equation}

If we omit the second order derivative of Eq. (\ref{reg-eqn}),   the loss function
is reduced to the
first-order Tikhonov regularizer.

With the generalized linear network assumption only for Jacobin regularization term, $ \sum _{i=1}^ N \|g^{\prime } (\mathbf{x}_i, \mathbf{W})\| ^2 =N  \sum _{j=1}^ M w_j ^2$, 
weight decay regularizer is obtained.

When we just simple let $\mathbf{z}$ is reconstructed input vector $\mathbf{x}$,  this will reduce into  the contractive auto-encoders loss function\cite{GYC2016}\cite{Rifai2011ContraAE}\cite{Rifai2011HiOCAE}.

\subsection{Numerical Method for Reaction-Diffusion Equation}

For non-equilibrium statistics, such as dissipative structures, we used the reaction-diffusion equation to study the dynamic process of the SLS. When studying a differential dynamical system, we are concerned with the properties (mainly global properties) of the system and its changes during perturbation. In a cooperative learning systems, the reaction-diffusion equation can also be used to describe the evolution dynamics of the system. If the SLS is designed as a differential dynamical system and attention is paid to the attractor subnetwork \cite{RollsComputational21}\cite{DomnisoruMembrane22}, the reaction-diffusion equation can also be used to describe it. Therefore, the mathematical basis of artificial intelligence should also include ordinary and partial differential equations.

Example: Turing's reaction-diffusion equations \cite{Turing1952}.

Turing's reaction-diffusion equations is one of his revolutionary discoveries in the field of natural science and is the mechanism for Pattern Formation\cite{Pearson1993}.

\begin{eqnarray}
\label{turing-equations}
 \frac{\partial U}{\partial t} &=& D_u \nabla ^2 U + f (U, V) \nonumber \\ 
 \frac{\partial V}{\partial t} &=& D_v \nabla ^2 V + g (U, V).
\end{eqnarray}

 It may be the  reason that inspired by Turing's discovery of reaction-diffusion equations,  Prigogine proposed the theory of the Dissipative Systems, believing that ``the energy exchange with the outside world is the fundamental reason for making the system orderly (contrary to the principle of entropy increase),'' and founded a new discipline called  non-equilibrium statistical mechanics \cite{Self-Organization26}.

{\bf Explanation}: Here, we interpret the two substances in Turing's model as two types of informations. The reaction of substances is similar to the processes of information fusion and production, and the diffusion of substances is equivalent to the process of information transmission. By modeling the SLS  as a processing system of information, the so-called Information Granular processing systems, we will study the general systems of artificial intelligence with this basic concept.  However, the information is not only photons like ``particles'', but also has ``wave-particle duality''.  In our SLS theory, information particles are described by high-dimensional random variables, and the function for information waves is the density distribution function.

{\bf Conjecture}: Turing's reaction-diffusion equations can be applied to pattern generation,  can we produce the topological structure of neural network with Turing's equations?  At here  our conjecture is: there exist  functions $f (U, V)$ and $g (U, V)$,   the neural network structure can be automatically generated by Turing's equation. The $U$ and $V$ stand for two different graph networks, respectively. (More discussions about this conjecture  will be further planned work  in  part III of the SLS.)

An information wave is different from an electromagnetic wave, which is the carrier of information, but the elliptic and parabolic equations in the mathematical physics equation can be used to describe the process of information transmission. Methods to solve such equations depend on the complexity  of the problem. At present, most studies on partial differential equations in mathematics use the finite element method \cite{Zienkiewicz2013} and the finite difference method \cite{FerzigerFinite28} to obtain numerical solutions. In our earlier work, we used the heat diffusion equation to study the propagation of light beams in nonlinear media \cite{Guo1990} \cite{Guo1990b} and the dynamics of interference filters \cite{GuoDynamics31}. The diffusion equation is a parabolic semi-linear partial differential equation and can also be used to study dispersive optical tomography \cite{Niu2008Improving32}.

In practice, uncertainty problems usually are tackled with Bayesian varies methods.  If we transform the uncertainty problems into a deterministic system learning problem, we can use the  stochastic gradient descent algorithm or a pseudoinverse learning algorithm to optimize the SLS\cite{guo1995exact} \cite{guo2001pseudoinverse} \cite{guo2004pseudoinverse}.  In our 2003 paper \cite{Guo2003reg}, we set one model as a parametric model and the other as a nonparametric model based on the two models. After variational inference, a second order approximation was adopted to turn it into a deterministic system learning (optimization) problem. The method is called approximate  synergetic learning algorithm as presented in subsection (\ref{sec-ASLA}).

\section{Discussion}

This paper briefly introduces the SLS's  concept, architecture and algorithms. Its main goals are to build a grand unified framework of the world model and to explore the road toward ``intelligent mechanics.''

However, by constructing the SLS,  can we achieve ``intelligent mechanics'' and develop grand unified theories about artificial intelligence?  We already know the significance of building a world model, but why study the grand unified theories?

The world model described by Yann LeCun is the world model of artificial intelligence, but the world we live in is a physical world. The intellectual activity of human beings belongs to the mental world. Therefore, the transition space when constructing an intelligence world model is the Cyber-Physics Model (CPM), which means that one of the vital cornerstones of artificial intelligence is physics, and physicists like to unify  these theories. In these theories,  complex phenomena are described as a set of concepts, and mathematical formulas that express these concepts can make very successful predictions.

In physics, the grand unified theories, super symmetry, and the M theory are not only very beautiful thoughts but also deal with many problems that cannot be solved by standard models, thus attracting many theoretical physicists. However, no matter how wonderful they are, they will eventually require extensive experimental verification. We need to remember that a good scientific theory must meet the following three conditions:

\begin{enumerate}
\item It must be able to replicate all successful predictions of existing scientific theories;
\item It should be able to explain the latest experimental and observational data that existing scientific theories cannot explain;
\item And most importantly, it should also make predictions that can be tested.
\end{enumerate}

Maxwell's equations, General Relativity, and the Standard Model all conform to these three points.

Therefore, as a theory, it must not only explain the present but also can predict the future. Does our SLS can meet these three points? We believe that the SLS basically meets these three points in the artificial intelligence world.

In physics, a theory needs to make predictions that can be tested, but in the field of artificial intelligence, a theory ought to be subversive and innovative. The innovation behind our synergetic learning theory is based on the principle of least action, the variation of free energy, and the reaction-diffusion equation. These equations describe the process of systems evolution. During the evolution of the systems, the systems is considered as a differential dynamics systems, which is described mathematically by the reaction-diffusion equations. Therefore, the intelligent ``thermodynamic system'' can be seen as the processing of information particles. The main innovation is that in the SLS there are at least  two subsystems, one subsystem is generated by the differential dynamic system and the other is determined by the reduction model.  We used the differential dynamic system to present the generative model, and  the reduction model is corresponds to the disentangle  model.

In systems science, a common phrase is ``complex world, simple rules.'' One of Yann LeCun' s questions is whether or not there is a simple rule behind learning. If we think that if there is such a simple rule, then what is it? At present, no one has answered this question yet, but we did it ! We found this rule: the principle of least action.

Why the principle of least action is the mentioned simple rule? It is a common sense that brain development is the Darwin process of ``Evolution \& Selection.''  The evolution of the human brain is not only related to the brain: it is also the sum of human evolutionary results and the coevolution of Earth's entire biological community. The study of bio-intelligence has never been focused a single individual but, rather, on the evolution of all organisms in all populations in the history of the world, and it is a learning process with the function of survival as the optimization goal. The objective function of natural selection is driven by the probability of survival, and in nature, it always prefers the state with the least energy. The objective function in machine learning is set by human beings, which is a choice by human beings and conforms to their laws. Therefore, the principle of least action is the law we chose for the evolution of the artificial intelligence world model.

The novelty of the SLS theory we propose is that most of the other methods consider either cooperation or competition; we believe that cooperation and competition exist among groups in a systems that is harmonious and coexisting, and this systems is the opposites unity of contradictory. During the process of evolution, the relationship among groups is not static but fluctuates from cooperation to competition and from competition to cooperation. In a Synergetic Learning Systems, competition is also a synergy.

\section{Summary and Perspectives}

In this work a method for addressing the challenge of difficult in AI research fields is proposed. We believe that under a given environment (data, boundary conditions), the solution to the differential equation can be predicted with a defined evolutionary path, and the final effect is predictable and controllable. However, the problem with uncertainty is that during the evolution of the systems from simple toward complex, a phenomenon that needs to pay attention is emerging. The term ``emergent phenomena,'' as used by condensed matter physicists, refers to the complex nature produced by the interaction between a large number of simple components. In life, the current phenomenon is the interaction between molecules and how the molecules combine to form a structure or perform a function. Living systems evolve, adapt and change through interactions or information exchanges with other systems. Biological systems have feedback loops that make them difficult to analyze using standard differential equations. We do not know how to solve this problem. Ramin Golestanian, director of the Max Planck Institute for Dynamics and Self-Organization, said: ``Physicists have studied many complex systems, but in terms of the number of complexity and degrees of freedom, life systems belong to a completely different category.''  Therefore, currently the SLS is not concerned to be a living systems, in fact it is  an artificial intelligence systems. How to integrate the emerging phenomenon into the differential dynamic equation during the evolution process is a future research direction.

The further work will focus on exploring the difficult mechanism explanation problem and the interpretable neural network model problem. The further work is aimed at addressing the challenges of topology design and explores the design of neural network topology.

The second part (part II) of SLS  is to explore the interpretable neural network model for the challenge of explainable mechanism of deep neural network model based on statistical mechanics  \cite{Synergetic2019b}. One scheme of interpretability is to interpret information processing as a process in which information is transformed through a complex system. Based on the big bang theory, the multi-models SLS interpretation holds that the systems is in the ground state at the beginning, driven by fluctuations and the force outside systems. And this system diffuses to the current state through long-term evolution. In general evolutionary computation, the goal of evolution is unknown. But in our SLS, the goal of evolution is approach to minimum of system's free energy.  The clustering problem (unsupervised learning) can be explained as the process of system reduction. Through the ``Maxwell's Demon'', the information particles gradually agglomerate from disorder to order state. 

The third part (part III) of SLS is to explore the topologic architecture design of neural network in view of the challenge of automatic machine learning. Based on the theory of system self-organization, the theory and method of network structure automatic organization and evolution will be developed.
In systems science, the theory of system self-organization studies how a systems automatically changes from disorder to order state, or from low-level order to high-level order state under certain conditions,  one example is the theory of laser.  From {\bf the thermodynamic point of view}, ``self-organization'' refers to a systems 
through the exchange of material, energy and information with the outside systems, the entropy of the system is constantly reduced and its order degree is improved. From {\bf the point of view of statistical mechanics}, ``self-organization'' refers to the spontaneous migration of a system along the  direction  from the most probable state to  lower probable state. 
From {\bf the point of view of evolutionism}, ``self-organization'' refers to a process in which a system under the influence of ``inheritance'', ``variation'' and ``survival of the fittest'', constantly improves its organizational structure and operation mode, so as to improve its adaptability to the environment.

\section*{Acknowledgement}

The research work described in this paper was fully supported  by the National Key Research and Development Program of China (No. 2018AAA0100203).  Prof. Ping Guo and Qian Yin are the authors to whom all correspondence should be addressed.



\end{document}